\documentclass[letterpaper, 10 pt, conference]{ieeeconf}  

\IEEEoverridecommandlockouts                              

\usepackage{algorithm}
\usepackage{algpseudocode}
\usepackage{graphicx}

\usepackage[dvipsnames]{xcolor}
\usepackage{graphics}
\usepackage{amsmath}

\usepackage{amssymb}
\usepackage{epsfig} 
\usepackage{booktabs}
\usepackage{float}
\usepackage{url}
\usepackage{gensymb}
\usepackage[shortcuts]{extdash}
\usepackage{xspace}

\newcount\Comments 
\usepackage{color}
\definecolor{darkgreen}{rgb}{0,0.5,0}
\definecolor{purple}{rgb}{1,0,1}
\newcommand{\kibitz}[2]{\ifnum\Comments=1\textcolor{#1}{#2}\fi}

\title{\LARGE \bf

Reciprocal Learning of Intent Inferral\\with Augmented Visual Feedback for Stroke

}

\author{Jingxi Xu$^{*, 1}$, Ava Chen$^{*, 2}$, Lauren Winterbottom$^3$, Joaquin Palacios$^{2}$, Preethika Chivukula$^{4}$,\\Dawn M. Nilsen$^{3,5}$, Joel Stein$^{3,5}$, and Matei Ciocarlie$^{2,5}$
\thanks{This work was supported in part by the National Institutes of Health: National Institute of Neurological Disorders and Stroke under grant R01NS115652, and Eunice Kennedy Shriver National Institute of Child Health and Human Development under award F31HD111301.}%
\thanks{$^*$These authors have contributed equally to this work.}%
\thanks{$^{1}$J. Xu is with the Department of Computer Science, Columbia University, New York, NY 10027, USA.
        {\tt\small jxu@cs.columbia.edu}}%
\thanks{$^{2}$A. Chen, J. Palacios, and M. Ciocarlie are with the Department of Mechanical Engineering,
        Columbia University, New York, NY 10027, USA.
        {\tt\small \{ava.chen, matei.ciocarlie\}@columbia.edu}}%
\thanks{$^{3}$L. Winterbottom, D. M. Nilsen, and J. Stein are with the Department of Rehabilitation and Regenerative Medicine, Columbia University Irving Medical Center, New York, NY 10032, USA.}%
\thanks{$^{4}$P. Chivukula is with the Department of Biomedical Engineering,
        Columbia University, New York, NY 10027, USA.
        }%
\thanks{$^5$ Co-Principal Investigators}
}

\begin{document}

\maketitle
\thispagestyle{empty}
\pagestyle{empty}

\begin{abstract}
Intent inferral, the process by which a robotic device predicts a user's intent from biosignals, offers an effective and intuitive way to control wearable robots.
Classical intent inferral methods treat biosignal inputs as unidirectional ground truths for training machine learning models, where the internal state of the model is not directly observable by the user. In this work, we propose reciprocal learning, a bidirectional paradigm that facilitates human adaptation to an intent inferral classifier. 
Our paradigm consists of iterative, interwoven stages that alternate between updating machine learning models and guiding human adaptation with the use of augmented visual feedback.
We demonstrate this paradigm in the context of controlling a robotic hand orthosis for stroke, where the device predicts \textit{open}, \textit{close}, and \textit{relax} intents from electromyographic (EMG) signals and provides appropriate assistance.
We use LED progress-bar displays to communicate to the user the predicted probabilities for \textit{open} and \textit{close} intents by the classifier. Our experiments with stroke subjects show reciprocal learning improving performance in a subset of subjects (two out of five) without negatively impacting performance on the others. We hypothesize that, during reciprocal learning, subjects can learn to reproduce more distinguishable muscle activation patterns and generate more separable biosignals. 

\end{abstract}
\section{Introduction}

Developing intuitive, user-driven methods for interfacing robotic devices with humans offers the potential to restore or enhance movement capabilities that are lost after neuromotor injuries such as stroke. Approaches that infer intent with the use of electromyography (EMG) and other biosignal data can encourage users to practice integrating the impaired limb in daily activities~\cite{Luo2024intent}, which in turn can facilitate neuroplasticity to improve motor function~\cite{Ballester2016nonuse}. 

Recent advances in machine learning (ML) can infer intent from EMG even when the strength of muscle activation is insufficient to move the limb~\cite{meeker2017emg, xu2022adaptive, Zhang2012pattern}. However, neuromotor impairments after stroke include abnormal muscle coactivation and diminished capacity to generate consistent EMG patterns. Such abnormalities in biosignal generation increase with volitional effort when attempting to use the arm~\cite{Chen2024volitional, Barry2022characteristics}, which in turn increases the challenge of intent inferral as a control method for robot-assisted movement.

\begin{figure}[t]
    \centering
    \includegraphics[width=1\linewidth]{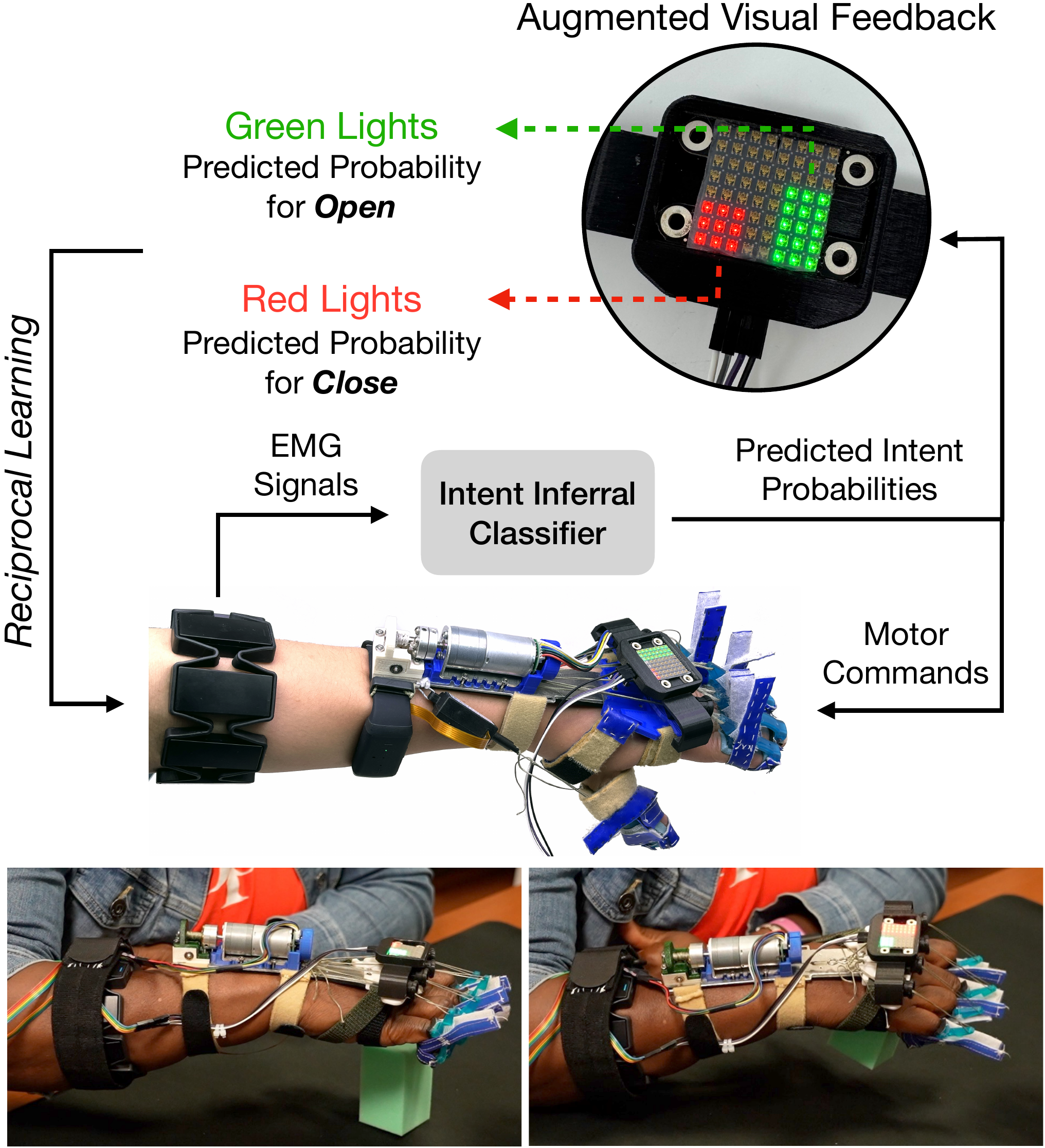}
    \caption{\textbf{Reciprocal learning with robotic hand orthosis and augmented visual feedback.} Our hand orthosis assists stroke survivors in opening the hand when an \textit{open} intent is detected from EMG signals, and allows the hand to close when it detects the \textit{close} intent. Augmented feedback consists of LED display bars corresponding to the predicted probabilities of \textit{open} and \textit{close} from the intent classifier running on the orthosis. Subjects use visual feedback to adapt to the intent inferral model during reciprocal learning practice. The bottom row shows a stroke survivor using our device with augmented feedback for a functional pick-and-place task.
    }
    \label{fig:teaser}
    \vspace{-4mm}
\end{figure}

Typical intent inferral methods take in user intents as unidirectional ground truths, but wearable interfaces are inherently bidirectional systems. Humans modify their control inputs in response to sensory feedback of device behavior, such as whether the device moves according to expectations~\cite{Proulx2022interaction}. Device developers have sought to provide augmented feedback, or feedback in addition to intrinsic human sensing, to communicate information about biosignals that are not directly observable in order to aid human adaptation. Augmented feedback can compensate for reduced proprioception associated with sensory impairments after stroke~\cite{Ingemanson2019somatosensory}, and when applied to EMG signals can offer insights into muscle coactivations~\cite{Wright2014mci}. Multiple groups~\cite{Madduri2023coadaptation, Hu2023coadaptation} have suggested augmented feedback of biosignals as a potential modality to mediate co-adaptive learning and enable personalized control, such as human and device simultaneously adjusting control parameters in response to the other agent's behavior. However, co-adaptation in practical deployments such as physically assistive devices or rehabilitative interventions has yet to be realized for non-able-bodied users.

Our approach uses bidirectional intent inferral on a robotic device to dynamically encourage greater separability of control input patterns by stroke-impaired users, instead of solely relying on static thresholding of physiological or kinematic signals~\cite{cisnal2023interaction}. Furthermore, we can immediately deploy this approach for robot-assisted movement. We treat the human as a dynamic co-learner alongside the classifier algorithm, prompting each to update their understanding of the other's behavior. The combination of an adaptive classification algorithm and augmented feedback can increase the coherence of sensory input and action output, which in turn helps users develop greater automaticity when controlling the hand. In reciprocal learning, the user learns to generate better biosignals for ML models while practicing actively using the hand. In summary, our main contributions are as follows:

\begin{itemize}
    \item We introduce a novel paradigm for training EMG-based intent inferral, consisting of interwoven sessions that alternate between updating ML models based on human biosignal generation and encouraging human exploration and adaptation to models with the use of augmented feedback.
    \item As far as we know, this bidirectional training system is a completely unexplored area in the context of upper-limb rehabilitation after stroke. We are the first to use augmented feedback to communicate the inner state of the learning model running on a wearable robotic orthosis as it assists hand movement.
    \item Our results demonstrate that reciprocal learning leads to better predictions for intent inferral on a subset of stroke subjects by aiding in the generation of more separable muscle activation patterns and biosignals. 
\end{itemize}
\section{Related Work}

\paragraph{EMG Intent Inferral for Stroke} ML offers a data-driven approach to learn personalized human motor pattern classification. Many works have demonstrated robust EMG control of exoskeletons by healthy users and amputees~\cite{Luo2024intent, Madduri2023coadaptation, Hu2023coadaptation}; however, the interpretation of biosignals from stroke-impaired users is a challenge. The majority of works on intent inferral for stroke use supervised~\cite{meeker2017emg,lee2010subject} or semi-supervised~\cite{xu2022adaptive} learning approaches to train models on an initial dataset for later use during that same session, but these paradigms are vulnerable to severe shifts in distribution due to scarce data and time available for training. Recent works have sought to generalize intent inferral to enable few-shot transfer from data trained on a cohort of stroke survivors~\cite{la2024meta, Sarwat2024fewshot}, or have attempted to expand the data available from a single training session with a given stroke subject by use of synthetic data generation~\cite{Xu2024chat}. However, these methods expect muscle activation inputs to not substantially change as the user attempts to repeat movements. Impaired users who are unable to execute, and therefore are unable to observe, movements due to muscle weakness after stroke generate highly variable and inconsistent EMG patterns~\cite{lee2010subject} that are prone to drift.

\paragraph{Interactive Learning with Augmented Feedback} Augmented feedback is a method of providing the user with information on biosignals by sources outside the body. Visual augmented feedback, including computer monitor displays or indication lights embedded on a device, has extensive support for improving rates of human motor adaptation beyond performing the task alone~\cite{Wang2024biofeedback}, and has long been deployed in EMG-based rehabilitation regimens for stroke~\cite{Wolf1983biofeedback}. Providing continuous, real-time concurrent feedback with human actions has widely shown benefit for stroke-impaired users to interactively learn mappings between different movement behaviors and resulting muscle activation patterns~\cite{Jian2021mci, Rinaldi2020alter}. In these works, augmented feedback is used to benchmark human efforts against a fixed healthy ideal; instead, we use augmented feedback to update the user about personalized device adaptation. Also, previous works focused on reinforcing or modulating already-learned activation-movement mappings, whereas we focus on the task of training a stroke-impaired user to operate a robotic orthosis that provides movement capabilities that the user does not already possess.

Recent works~\cite{Madduri2023coadaptation} have combined augmented feedback with task-specific training to develop new control inputs. Seo et al.~\cite{Seo2023healthy} altered flexor synergies in healthy subjects using EMG signal-guided conditioning on an isometric task. Silversmith et al.~\cite{Silversmith2021bci} used co-adaptation to enable a paralyzed subject to add additional control dimensions to a brain-computer interface. However, our reciprocal learning paradigm uniquely co-trains adaptation in the context of physical devices and movement. To our knowledge, no previous works have tackled the challenge of mediating intents between adaptive devices and adaptive users. 
\section{Reciprocal Learning Paradigm}

\begin{figure*}[h]
    \centering
    \includegraphics[width=0.9\linewidth]{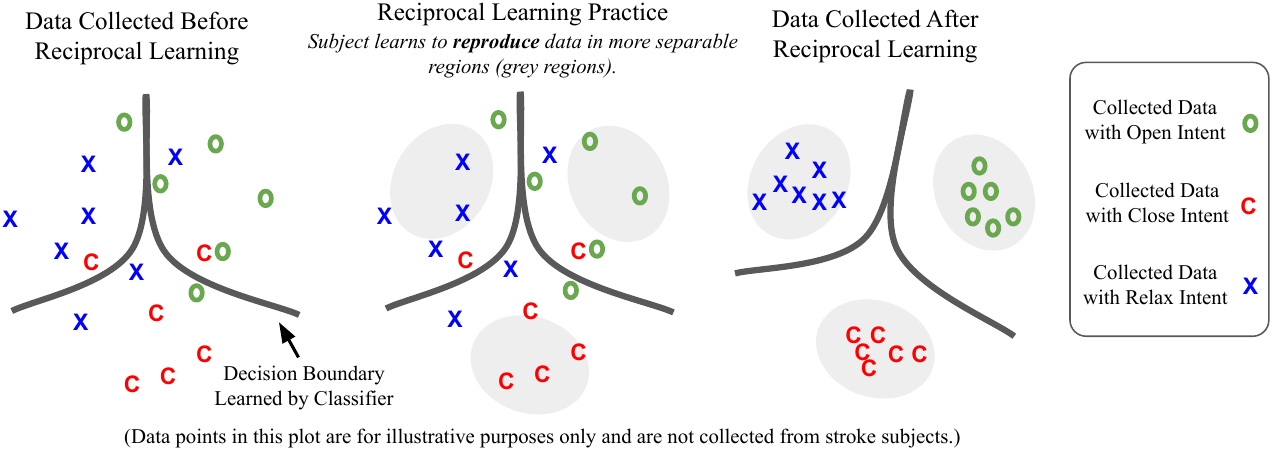}
    \caption{\textbf{The goal of reciprocal learning is to guide the subject in generating more separable signals.} Before reciprocal learning practice, the collected data are more diverse for each intent, making it difficult for the classifier to learn a perfect decision boundary. During reciprocal learning practice, the subject adapts to the intent inferral classifier and learns to generate data in more separable regions. After reciprocal learning practice, we retrain the classifier using the newly generated more separable data. This figure is intended to illustrate the concept, and does not show EMG signals collected from real subjects; separability data from real stroke subjects will be shown in Fig.~\ref{fig:separability}.}
    \label{fig:illustration}
\end{figure*}

\begin{figure*}[h]
    \centering
    \includegraphics[width=1\linewidth]{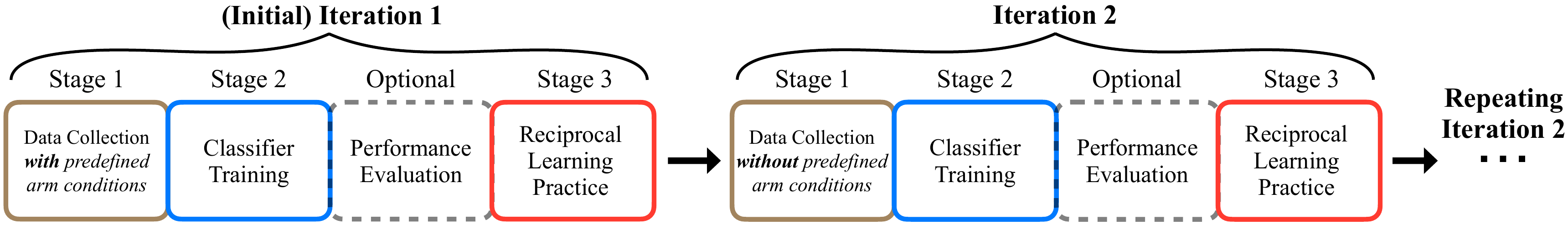}
    \caption{\textbf{Reciprocal learning paradigm.} An iteration consists of data collection, classifier training, performance evaluation, and reciprocal learning practice.}
    \label{fig:paradigm}
    \vspace{-2mm}
\end{figure*}

The ultimate goal of this project is to develop algorithms that can infer intent (one of \{\textit{open}, \textit{relax}, \textit{close}\}) from EMG signals generated by stroke-impaired users attempting to move their hands, such that our robotic device can provide appropriate assistance. If the predicted intent is \textit{open}, the robot extends the user's fingers to open the hand; if the predicted intent is \textit{close}, the robot releases to allow the user to use their own strength to close the hand; if the predicted intent is \textit{relax}, the robot simply maintains the previous state. 

The classical way to train such intent inferral algorithms is completely unidirectional. A researcher would issue cues to open, close, or relax the hand to a user wearing the device, then would record the real-time EMG signals and label the data using the provided cues as ground truth intent. In this unidirectional process, ML models learn personalized patterns through the labeled dataset and generate predictions. Although the user can observe the robot opening or closing, they do not have knowledge about the underlying ML predictions or EMG signals that led to the robot taking action.

Our approach aims for a different bidirectional paradigm, dubbed reciprocal learning. We hypothesize that by providing insights about the internal state of the model in an intuitive and self-explanatory way, a user can adapt to the behavior of the classifier and reproduce more distinguishable control inputs. This idea is illustrated conceptually in Fig.~\ref{fig:illustration}.


Reciprocal learning is an iterative process and can have a potentially unlimited number of iterations (shown in Fig.~\ref{fig:paradigm}). Each iteration consists of multiple stages, listed below: 

\begin{itemize}
    \item \textbf{Stage 1: Data Collection.} We collect a labeled dataset from a subject by providing verbal cues of \textit{open}, \textit{relax} and \textit{close} while simultaneously recording EMG signals.  
    \item \textbf{Stage 2: Classifier Training.} We train an intent inferral classifier using the collected dataset. This is where one direction of the learning happens---the ML model learns from the subject through the labeled dataset. 
    \item \textbf{Optional: Performance Evaluation.} The purpose of this optional stage is to use any metric of our choice to track the performance of the trained intent inferral classifiers across iterations. 
    \item \textbf{Stage 3: Reciprocal Learning Practice.} We run the learned intent inferral classifiers on the orthosis and provide the augmented visual feedback to communicate the inner state of the ML model that is not directly observable. The subject practices to generate distinguishable biosignals for \textit{open} and \textit{close} intents, using augmented feedback as guidance. This is where the opposite direction of learning happens---the subject explores and adapts to the ML model.
    \item \textbf{Repeat...} We continue to the next iteration, and collect a new batch of data from the subject who has supposedly learned to produce more distinguishable EMG signals.
\end{itemize}

\section{Methods and Implementation}

Having described our general concept and approach, we now expand on the method and implementation of each stage.

\subsection{Data Collection}
In the data collection stage of each iteration, we collect a labeled dataset by providing verbal cues for subjects to open, close, and relax the hand. We simultaneously record the EMG signals and the given verbal cues as ground truth intent labels. The EMG armband \mbox{(Myo, Thalmic Labs)} has eight electrodes encircling the forearm and samples 8-channel EMG signals from the forearm surface at 50Hz. 

Note that the arm conditions under which the data is collected differ between the first iteration and the following iterations. A \textit{condition} is a particular combination of hand position and whether the orthosis is actively providing assistance. In the initial iteration, the orthosis has no existing classifiers to run and thus no augmented visual feedback (see Sec.~\ref{sec:feedback} for details on the feedback mechanism) to provide. The subject has not done any reciprocal learning practice and thus has not found any muscle activations that consistently generate distinguishable EMG signals. As a result, we collect data from a diverse set of predefined arm conditions in the hope that this set provides enough diversity of muscle activation patterns such that one of them or interpolation between them is easily reproducible and separable. 

The four different arm conditions of the initial iteration are as follows. (1) \textit{Arm on table, motor off}: the assisted arm of the subject rests on the table, and the motor of the orthosis is disengaged, providing no physical assistance to the subject. (2) \textit{Arm on table, motor on}: the assisted arm of the subject rests on the table, and the motor of the orthosis is engaged, allowing the device to provide active grasp assistance to the user. (3) \textit{Arm off table, motor off}: the subject keeps their arm raised to shoulder level for the duration of data collection, and the motor is disengaged, providing no physical assistance to the subject. (4) \textit{Arm off table, motor on}: the subject keeps their arm raised to shoulder level for the duration of data collection, and the motor of the orthosis is engaged, allowing the device to provide active hand-opening assistance.

For each condition, we collect two recordings, one for training and the other one held out for testing. A \textit{recording} is defined to be a continuous and uninterrupted recorded sequence of EMG signals. For each recording, we instruct the subject to open and close their hand 3 times. Each verbal cue lasts for 5 seconds, with a \textit{relax} cue between each \textit{open} and \textit{close} cue. For conditions where the motor is active, the orthosis moves approximately 1 second after the verbal cue. 

In the data collection stage of the following iterations, there are no more predefined arm conditions. The orthosis runs the trained classifier, providing active assistance and augmented visual feedback. The subject's goal is to reproduce the muscle activation patterns learned in the reciprocal learning practice of the previous iteration. We collect four recordings: two for training and the other two for testing.

\subsection{Classifier Training}

In each classifier training stage, we train a Linear Discriminant Analysis (LDA) classifier. We choose LDA because it is fast, trackable, and widely used in biomedical research~\cite{meeker2017emg,xu2022adaptive,la2024meta,Xu2024chat}. In the optional performance evaluation stage, we evaluate the performance of the classifiers both qualitatively and quantitatively (see Sec.~\ref{sec:exp} for details).

The classifiers take in a single time step of an 8-channel, 50Hz EMG signal. We clip the signal to $[0, 1000]$ and then normalize it to $[-1, 1]$. Prediction on a single step tends to be noisy, so we apply a median filter of window size 20 to the predicted probabilities of \textit{relax}, \textit{open}, and \textit{close} over time. This median filter smooths out the predicted probabilities and removes unnecessary frequent changes in the intent.

\subsection{Reciprocal Learning Practice}
\label{sec:feedback}

Our augmented visual feedback consists of two progress-bar displays mounted on the dorsal side of the hand orthosis so that the subject can observe them comfortably during reciprocal learning, as shown in Fig.~\ref{fig:teaser}. The relative heights of the green and red LED bars correspond to the model's predicted probabilities for \textit{open} and \textit{close} intents, respectively.

In the reciprocal learning practice stage, we ask the subject to practice opening and closing their hands guided by the augmented visual feedback. We run the trained classifier on the device to provide appropriate assistance. In this stage, we instruct the subject to trigger different device states (\textit{open} and \textit{close}) by maximizing the respective green or red LED bars. 

This is the exploration stage, where the subject adapts to the behavior of the device, and tries to discover and reinforce the muscle activation pattern that is easily reproducible and distinguishable. In this stage, the subject uses augmented visual feedback to learn to generate more separable EMG signals for each intent. This idea is conceptually illustrated in Fig.~\ref{fig:illustration}. Reciprocal learning practice essentially helps the subject locate the EMG subspace that is both reproducible and separable (i.e., far from the \textit{decision boundary} learned by the classifier of the previous iteration). Then, in the next iteration, the subject can reproduce data in these more separable regions, leading to new retrained classifiers that can more accurately predict user intent. 


\begin{figure*}[t]
    \centering
    \includegraphics[width=\linewidth]{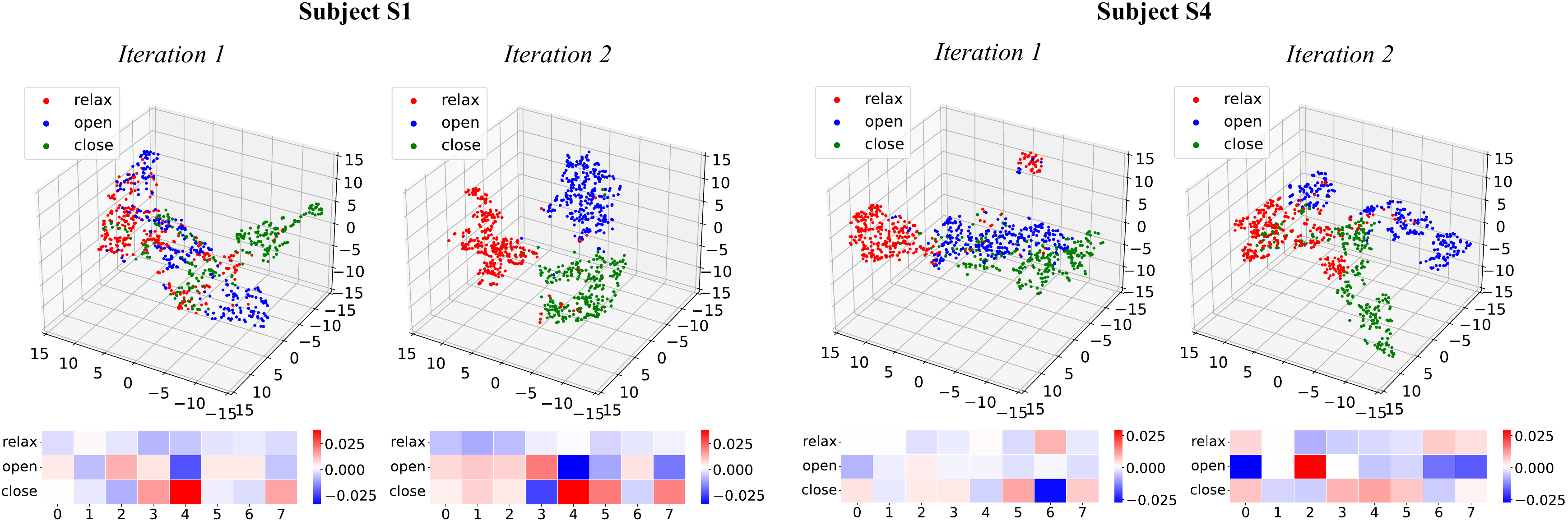}
    \caption{\textbf{Reciprocal learning results on two stroke subjects.} \textbf{Top:} comparing data separability. After reciprocal learning, clusters associated with each intent become more easily separable. \textbf{Bottom:} LDA classifier weights visualization indicating a specific EMG electrode (0 through 7) is negatively (blue) or positively (red) associated with a specific intent. Reciprocal learning strengthens such associations, which helps separability.}
    \label{fig:separability}
    \vspace{-2mm}
\end{figure*}

\section{Experimental Setup}

We tested reciprocal learning in single-session experiments with five stroke subjects, and compared intent inferral accuracies from two iterations. Reciprocal learning supports multiple iterations; however, here we only complete two iterations due to session time constraints. We leave the study of reciprocal learning with more iterations as future work.
\begin{table}[b]
\vspace{-2mm}
\centering
\caption{\textbf{Subject information.}}
 \begin{tabular}{c c c c} 
 \toprule
 Subject ID & Age & Gender & FM-UE \\
 \midrule
 S1 & 46 & Male & 27 \\ 
 S2 & 48 & Male & 26 \\
 S3 & 53 & Female & 26 \\
 S4 & 31 & Male & 50 \\
 S5 & 53 & Male & 47 \\ 
 \bottomrule
 \end{tabular}
 \label{tab:subjects}
 \vspace{-2mm}
\end{table}

\subsection{Subjects}

We performed experiments with five chronic stroke survivors having hemiparesis and moderate muscle tone: Modified Ashworth Scale (MAS) scores $\leq$ 2 in the upper extremity. Our MAS criteria exclude subjects whose fingers are difficult to move passively---fingers with more severe spasticity cannot be quickly extended with external forces without increasing muscle tone and risking damage to the joints. Our subjects can fully close their hands but are unable to completely extend their fingers without assistance. The passive range of motion in the fingers is within functional limits. Testing was approved by the Columbia University Institutional Review Board (IRB-AAAS8104) and was performed in a clinical setting under the supervision of an occupational therapist. See Table~\ref{tab:subjects} for details on subjects.

Our subjects have different degrees of hand impairment severity, measured by their varied Fugl-Meyer scores for upper extremity (FM-UE). Subjects S1, S2, and S3 have no active finger extension (minimal observable movement) and lower corresponding FM-UE scores (27, 26, 26, respectively), whereas S4 and S5 have some residual active finger extension capacity (limited observable movement) and higher corresponding FM-UE scores (50, 47, respectively). 

\subsection{Hardware}

Our hand orthosis~\cite{park2018multimodal,chen2022thumb} is a robotic exotendon device with a forearm EMG armband and LED visual displays, as shown in Fig.~\ref{fig:teaser}. The device assists hand opening by extending all digits simultaneously. It does not directly assist finger flexion and relies on human strength for hand closing.
\section{Results and Discussion}
\label{sec:exp}

In this section, we describe the evaluation of intent inferral performance across iterations and show the potential effectiveness of reciprocal learning. 

\subsection{Intent Inferral Accuracy}

\begin{table}[t]
\centering
\caption{\textbf{Intent inferral accuracy.}}
 \begin{tabular}{c c c c c c} 
 \toprule
 Subject & S1 & S2 & S3 & S4 & S5 \\
 \midrule
 Iteration 1 & 0.61 & 0.69 & \textbf{0.70} & 0.86 & \textbf{0.82} \\ 
 Iteration 2 & \textbf{0.88} & \textbf{0.71} & 0.68 & \textbf{0.94} & 0.80 \\
 \bottomrule
 \end{tabular}
 \label{tab:accuracy}
 \vspace{-2mm}
\end{table}

During the performance evaluation stage of each iteration, we compute the intent inferral accuracy of the trained classifier on the test set, shown in Table~\ref{tab:accuracy}. While the intent inferral accuracies for S2, S3, and S5 remain similar across the first two iterations, we observe an accuracy improvement for S1 and S4, from 0.61 to 0.88 and from 0.86 to 0.94, respectively. Augmented visual feedback during reciprocal learning practice may have helped S1 and S4 generate more consistent and distinguishable muscle activation patterns, creating more separable intent signals. 
As a result, the data collected after the first reciprocal learning practice become easier to classify. We note that S1 has more limited hand movement capacity compared to S4 (also reflected by their intent inferral accuracy). This suggests that whether reciprocal learning can improve intent inferral accuracy may not depend on stroke subjects' residual hand function. 

One possibility for why reciprocal learning does not improve performance across all subjects is the randomness of exploration in the practice stage. Although the prescribed arm conditions during the data collection stage of the first iteration provide data variance, due to the variation of impairment across stroke subjects, it is not guaranteed that the initial dataset is diverse enough to contain reproducible and distinguishable muscle activation patterns for all intents. In addition, even though the initial dataset collected with predefined arm conditions contains many activation patterns, due to randomness of exploration, it is not guaranteed for a subject to successfully locate or reproduce a specific pattern. 

While advancing the proposed paradigm to have a positive effect on most participants is obviously desirable and is the focus of our future work, we believe that an approach that improves performance on a subset of participants without hindering performance on the others can also have practical usefulness. Furthermore, a detailed study of the differences between cases showing successes versus more limited effects can suggest future areas of exploration.

\subsection{Data Separability}

\begin{table}[t]
\centering
\caption{\textbf{Variance of LDA weights for the open intent.}}
 \begin{tabular}{c c c c c c} 
 \toprule
 Subject & S1 & S2 & S3 & S4 & S5 \\
 \midrule
 Iteration 1 & $7.8\mathrm{e}{-6}$ & $1.0\mathrm{e}{-4}$ & $8.8\mathrm{e}{-4}$ & $1.1\mathrm{e}{-4}$ & $1.0\mathrm{e}{-4}$ \\ 
 Iteration 2 & $\mathbf{2.5\mathrm{\textbf{e}}{-4}}$ & $\mathbf{4.6\mathrm{\textbf{e}}{-4}}$ & $\mathbf{1.2\mathrm{\textbf{e}}{-3}}$ & $\mathbf{2.8\mathrm{\textbf{e}}{-4}}$ & $\mathbf{2.0\mathrm{\textbf{e}}{-4}}$ \\
 \bottomrule
 \end{tabular}
 \label{tab:variance}
 \vspace{-2mm}
\end{table} 

We further investigate the data of subjects S1 and S4 to show how reciprocal learning may lead to biosignals with more discriminative information for classifying intent. We visualize 1000 randomly sampled 8D EMG signals for each intent before and after the reciprocal learning practice in 3D space using t-distributed stochastic neighbor embedding \mbox{(t-SNE)}~\cite{van2008visualizing}, as shown in Fig.~\ref{fig:separability}. Each data point is labeled using its ground truth label, which is only available during training. After reciprocal learning, the data clusters associated with each intent improve in separability, which could explain the increased classification accuracy and lend support to the hypothesis that reciprocal learning may help some subjects generate signals that are more easily interpretable by intent inferral algorithms. 

We also visualize the LDA weights associated with each EMG electrode in Fig~\ref{fig:separability}. Each of the eight electrodes uniformly distributed on the forearm circumference covers a particular area of forearm muscles. Positive values (red) mean that muscle activity under that particular electrode is positively associated with a specific intent, while negative values (blue) mean that muscle activity contributes negatively to that intent. We notice that after reciprocal learning practice, the weights (especially for the \textit{open} intent) become more diverse, meaning that the positive weights are more positive, and the negative weights are more negative. In fact, the variance of the 8D weights for the \textit{open} intent increases consistently for all subjects, as shown in Table~\ref{tab:variance}. These results suggest that subjects have stronger muscle activation patterns after reciprocal learning practice and explain improvements in intent inferral accuracy from another angle.

\section{Conclusion}

We propose a novel bidirectional reciprocal learning paradigm that co-trains stroke-impaired users with intent inferral algorithms. We achieve reciprocal learning through augmented visual feedback in the form of LED progress-bar displays, which communicate the probabilities of the \textit{open} and \textit{close} intents predicted by the classifier. Our paradigm consists of iterative and interwoven stages of training the machine learning models and human exploration and adaptation to the classifiers. By using the augmented visual feedback of the classifier internal state as guidance, a subset of our stroke subjects learn to modulate muscle activation and generate more separable EMG signals, leading to better intent inferral performance, while the performance of the other subjects is not adversely affected. Future work includes conducting detailed studies of reciprocal learning involving multiple iterations, and exploring the factors that influence a subject's ability to find consistent and reproducible muscle activation patterns during reciprocal learning practice.


\bibliographystyle{IEEEtran} 
\bibliography{references} 

\end{document}